\documentclass[journal,comsoc]{IEEEtran}

\usepackage[T1]{fontenc}

\usepackage{ifpdf}
\usepackage{cite}
\usepackage{graphicx}
\graphicspath{{./Fig/}}
\usepackage{amsmath,amsfonts}
\usepackage{bm}
\usepackage{algorithm}
\usepackage{algorithmic}
\usepackage{array}
\usepackage[caption=false,font=normalsize,labelfont=sf,textfont=sf]{subfig}
\usepackage{url}
\begin{document}

\title{Study of MRI-compatible Notched Plastic Ultrasonic Stator with FEM Simulation and Holography Validation}

\author{Zhanyue~Zhao,
        Haimi~Tang,
        Paulo~Carvalho,
        Cosme~Furlong,
        and~Gregory~S.~Fischer% <-this % stops a space
\thanks{Z. Zhao, P. Carvalho, and G. S. Fischer are with the Department of Robotics Engineering, Worcester Polytechnic Institute, Worcester, MA 01605 USA e-mail: zzhao4@wpi.edu, gfischer@wpi.edu.}% <-this % stops a space
\thanks{H. Tang and C. Furlong are with the Department of Mechanical \& Materials Engineering, Worcester Polytechnic Institute, Worcester, MA 01605 USA.}% <-this % stops a space
\thanks{This research is supported by National Institute of Health (NIH) under the National Cancer Institute (NCI) under Grant R01CA166379.}
}

\markboth{IEEE Transactions on Ultrasonics, Ferroelectrics, and Frequency Control}%
{Shell \MakeLowercase{\textit{et al.}}: Bare Demo of IEEEtran.cls for IEEE Communications Society Journals}

\maketitle

\begin{abstract}

Intra-operative image guidance using magnetic resonance imaging (MRI) can significantly enhance the precision of surgical procedures, such as deep brain tumor ablation. However, the powerful magnetic fields and limited space within an MRI scanner require the use of robotic devices to aid surgeons. Piezoelectric motors are commonly utilized to drive these robots, with piezoelectric ultrasonic motors being particularly notable. These motors consist of a piezoelectric ring stator that is bonded to a rotor through frictional coupling. When the stator is excited at specific frequencies, it generates distinctive mode shapes with surface waves that exhibit both in-plane and out-of-plane displacement, leading to the rotation of the rotor. In this study, we continue our previous work and refine the motor design and performance, we combine finite element modeling (FEM) with stroboscopic and time-averaged digital holography to validate a further plastic-based ultrasonic motor with better rotary performance. 

\end{abstract}


\section{Introduction}

\IEEEPARstart{S}{urgeries} for cancerous tumor removal have a higher long-term success rate when adequate margins are achieved \cite{jacobs2008positive}. Intraoperatively guided procedures using magnetic resonance imaging (MRI) have been shown to achieve better margins compared to traditional methods \cite{senft2011intraoperative}. However, the strong magnetic fields, fast-changing gradients, sensitivity to electrical noise, and constrained space make operating within an MRI challenging. Robotic assistive devices can assuage these challenges \cite{futterer2010mri, monfaredi2018mri}. 

Conventional motors cannot be used inside an MRI environment. As a result, MRI-compatible robotic devices rely on alternative actuators, such as pneumatic \cite{stoianovici2007mri, fischer2008mri}, hydraulic \cite{kokes2009towards}, or piezoelectric actuators\cite{nycz2017mechanical}. However, in piezoelectric motors, non-ferromagnetic metals are still present, which can sometimes cause distortions in the magnetic field, affecting image quality. Previous research has shown that substituting several of these metal components with plastic equivalents can reduce magnetic field distortions without compromising the image signal-to-noise ratio \cite{carvalho2018demonstration}.

We conduct high-speed stroboscopic and time-averaged holography on a notched plastic stator to demonstrate its potential for constructing a more MRI-compatible resonant piezoelectric motor. The plastic stator design is based on a modified version from our previous work \cite{zhao2021preliminary,carvalho2020study}, where the dimensions were adjusted to closely match the frequencies of a commercial stator. We used a digital holographic system to validate the time-dependent FEM simulation and proved an improvement in rotary performance based on the initial flat plastic stator.

\section{Geometry Design}

\begin{figure}
    \centering
    \includegraphics[width=\linewidth]{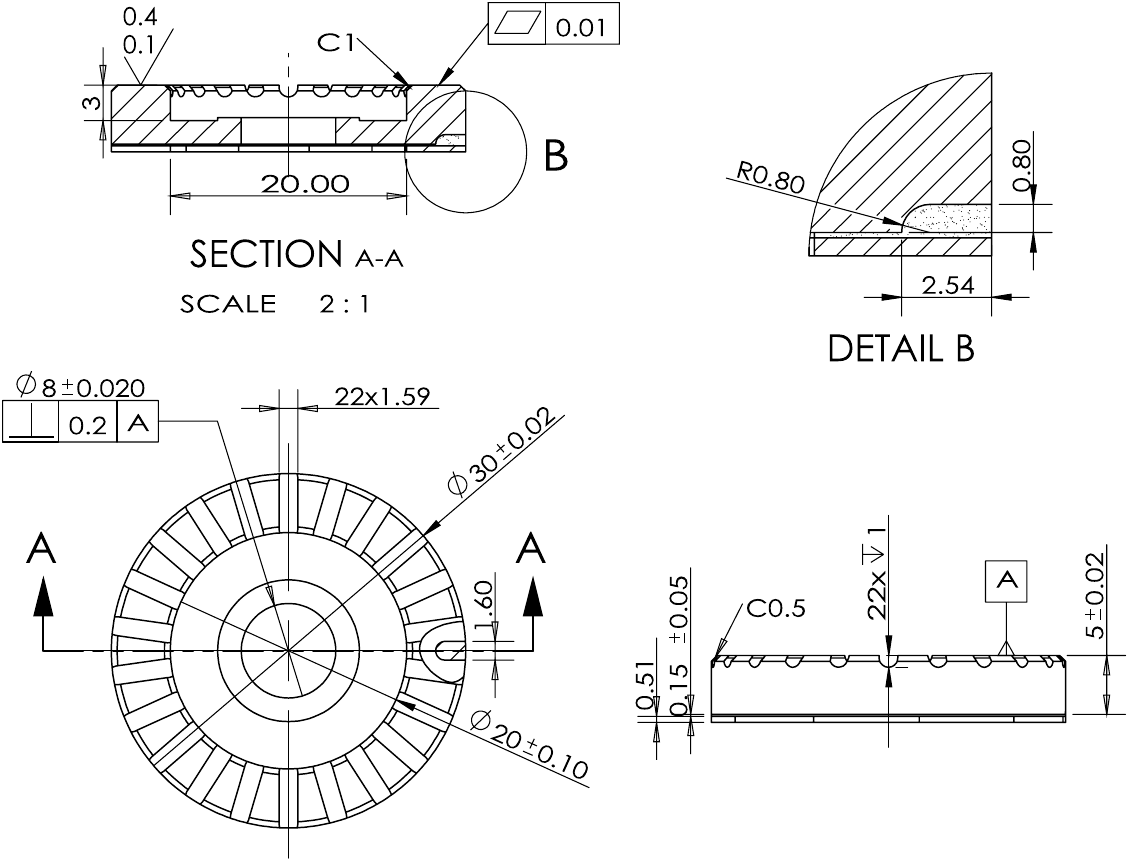}
    \caption{Dimensions of notched stator design, 22 notches with 1.59mm by 1mm were added to the stator based on the flat design.}
    \label{fig:notchgeometry}
\end{figure}

The notched stator used the same ceramic setup mentioned in \cite{zhao2021preliminary,carvalho2020study}, but some modifications on the plastic stator, which is shown in Fig. \ref{fig:notchgeometry}. Based on the flat stator design, 22 notches with 1.59mm width by 1mm depth were machined using 1/16in ball-end milling bit on a CNC machine. Other dimensions and tolerance requirements remained the same as the flat design.

\begin{figure}
    \centering
    \includegraphics[width=\linewidth]{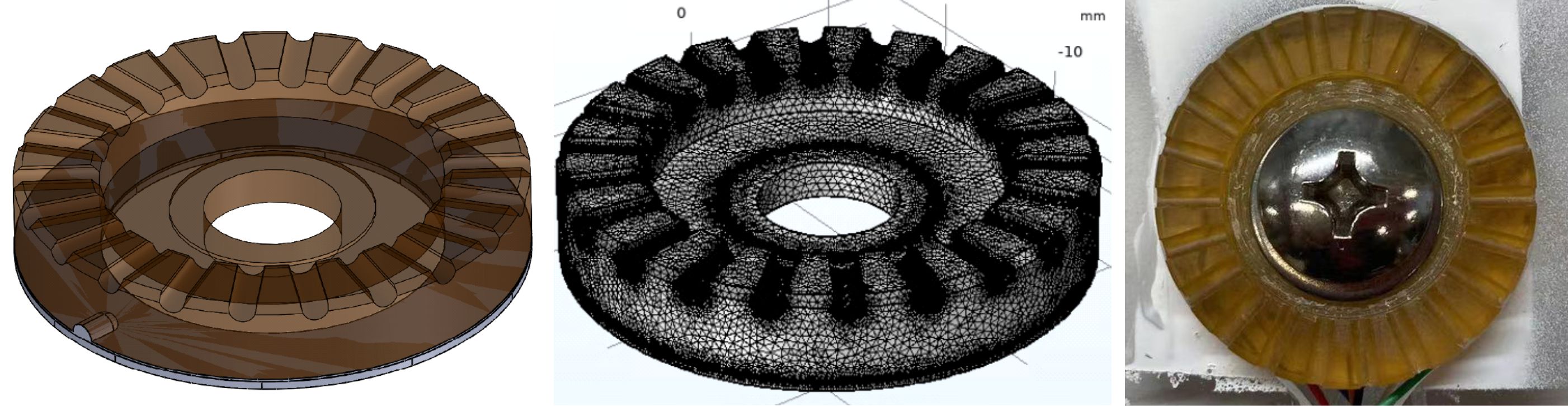}
    \caption{Notched stator design. (Left) CAD drawing of the notched stator. (Middle) FEM meshing of the notched stator. (Right) The notched stator made by Ultem plastic was fixed during the testing stage.}
    \label{fig:notch_design}
\end{figure}

Fig. \ref{fig:notch_design} shows the CAD drawings of the notched stator design. A FEM simulation with the same parameter and configuration discussed in \cite{carvalho_2020,zhao2023preliminary} was made and the middle figure shows the meshing configuration. Construction of a notched stator made of Ultem 1000 plastic is shown in the right figure, where the total height is 5.02mm and fixed onto a plastic testing stage. 

\section{Time-Dependent FEM Simulation}

\begin{figure}
    \centering
    \includegraphics[width=.9\linewidth]{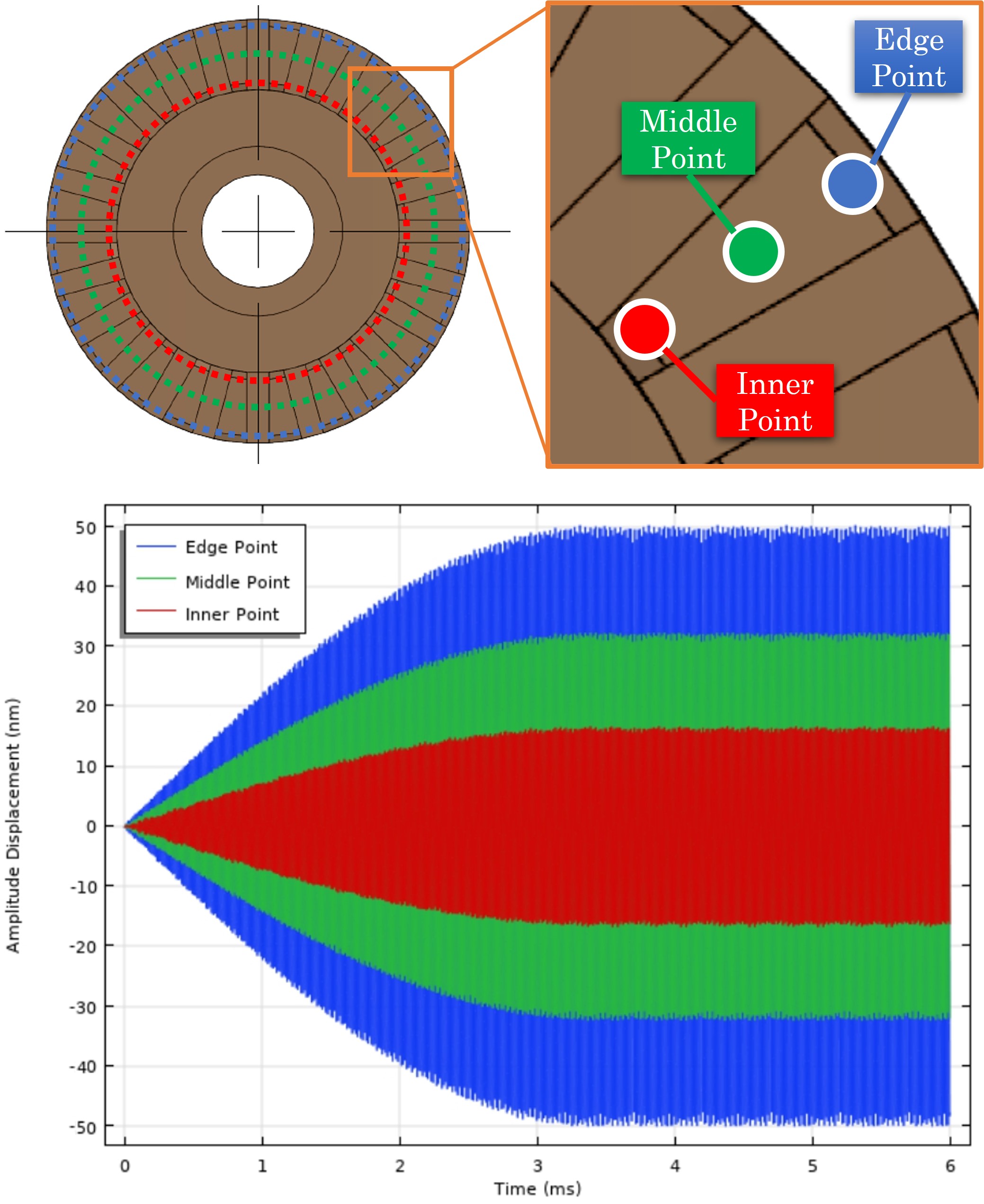}
    \caption{configuration and results of multiple point probe monitoring the out-of-plane displacement. The motor settling time is around 3.4ms, where the inner, middle, and edge circumference reaches the maximum amplitude around $\pm$30nm, $\pm$60nm, and $\pm$100nm respectively.}
    \label{fig:amp_plasticAll}
\end{figure}

A FEM time-dependent simulation was created for detailed displacement amplitude numerical prediction. Fig. \ref{fig:amp_plasticAll} shows the configuration and results of multiple point probes monitoring the out-of-plane displacement. We chose three points on one tooth's top surface, which was the contact surface between the stator and rotor as the research target, namely the inner point (red), middle point (green), and edge point (blue). The three points chosen represented the three circumference locations of the stator contacting surface. The right figure shows the numerical results of different point out-of-plane displacement, which indicates that under 100V peak-to-peak excitation voltage conditions, the motor settling time is around 3.4ms, where the inner, middle, and edge circumference reaches the maximum amplitude around $\pm$30nm, $\pm$60nm, and $\pm$100nm respectively. The increase of displacement amplitude from the inner to edge circumference was caused by the fixed constrain in the center lateral area discussed in our previous work \cite{carvalho_2020,zhao2023preliminary}, because the inner circumference is closer to the fixture area and the motion was reduced compared to the middle and edge circumference area. 

\section{Time-Averaged and Stroboscopic Holography}

\begin{figure}
    \centering
    \includegraphics[width=0.75\linewidth]{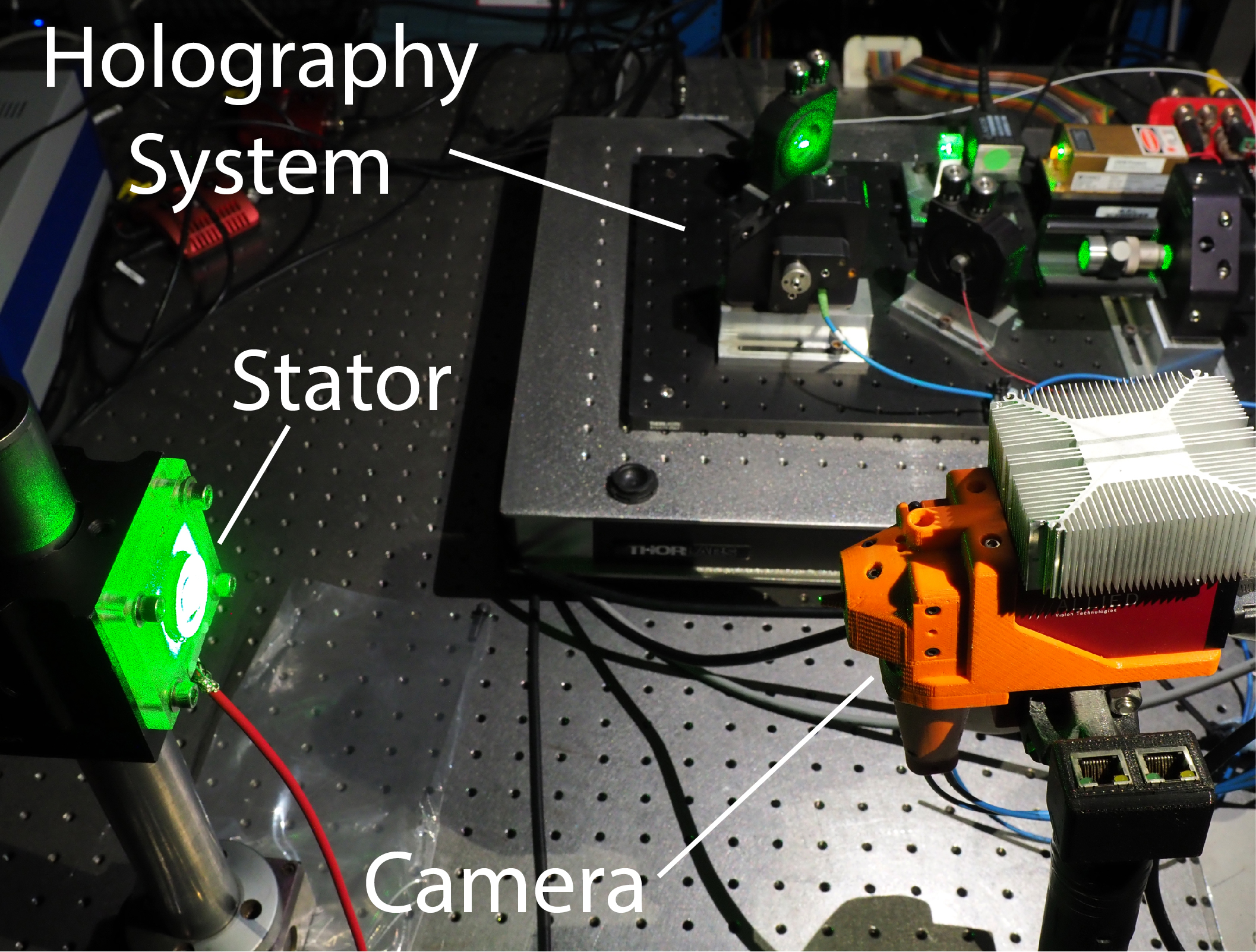}
    \caption{Plastic flat stator and digital holographic system on an optical table set up for quantitative imaging with nanometer resolution in full-field. Image imported from \cite{carvalho_2020}.}
    \label{fig:HoloSystem}
\end{figure}

In this study, a Digital Holographic System (DHS) was set up for stator imaging consisting of a laser delivery system with temporal phase stepping, stroboscopic illumination, and fiber-coupled output, a camera system (otoscope head) for lensless digital holography, and its supporting mechatronic otoscope positioner (MOP) which can be found in Fig. \ref{fig:HoloSystem}. The laser delivery system consisted of a 532nm 30mW laser (BWN-532-20E, B\&W TEK INC, USA) aimed through an acoustic shutter (AOM-40, IntraAction Corp, USA), the resulting beam is reflected by a mirror fixed to a piezo microactuator powered by a piezo controller (MDT694A, Thorlabs, USA). The camera system, or otoscope head (OH) consisted of a camera (AVT Pike F505B, Germany), integrated with an object illumination component providing 532nm laser light with 19$mW/cm^2$ intensity at the flat stator surface. This lensless DHS allows digital focusing post-acquisition and provides a 2M pixels resolution. The system is a variation of the one used in \cite{ebrahimian2021material}.

\begin{figure}
    \centering
    \includegraphics[width=1\linewidth]{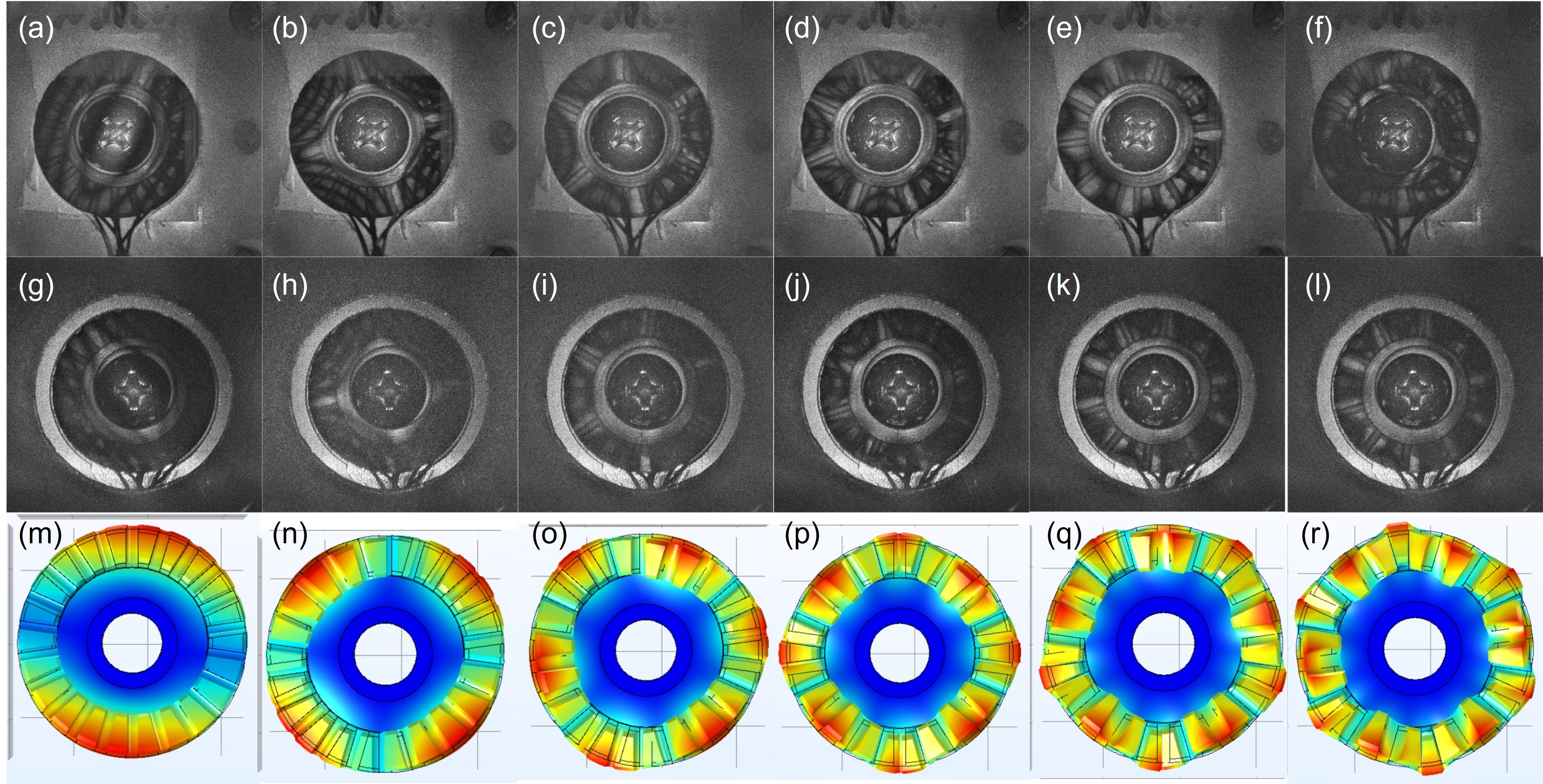}
    \caption{Notched plastic stator time-averaged excitation patterns versus simulation eigenmodes.}
    \label{fig:notch_holo}
\end{figure}

Fig. \ref{fig:notch_holo} shows the Notched plastic stator excitation frequencies from time-averaged results versus simulation eigenfrequencies results. The top and middle rows from (a) through (f) and (g) through (l) show two equivalent notched stators excitation patterns from 1$^{st}$ to 6$^{th}$ excitation mode respectively, and the last row from (m) through (r) represents the simulation results of 1$^{st}$ to 6$^{th}$ eigenmode patterns. Both results show the same motion pattern between the holographic images and the numerical results. Table. \ref{tbl:notched} shows the specific data of excitation frequencies and the eigenfrequencies from FEM simulation results. The difference between the simulation and experimental data is small (within 12.60\%). Note that one of the stators showed a mixed pattern in 6$^{th}$ excitation frequency so in the table is "-".

\begin{table}[ht!]
\caption{Notched plastic stator excitation frequencies from time-averaged results versus simulation eigenfrequencies results. Md - mode number, NPM - notched plastic motor, unit - kHz}
    \vspace{1mm}
    \centering
    \begin{tabular}{cccccccc}
    \hline
           & Md1  & Md2  & Md3  & Md4  & Md5  & Md6  & Md7  \\ \hline\hline
    NPM1       & 3.68   & 6.77   & 14.74  & 23.57  & 33.03  & -  & -    \\ \hline
    NPM2       & 3.97   & 6.98   & 14.50  & 23.08  & 31.86  & 42.63    & -    \\ \hline
    Simulation & 3.68 & 6.10 & 13.69 & 22.36 & 31.27 & 41.15 & 48.87 \\ \hline
    \end{tabular}
    \label{tbl:notched}
\end{table}

\begin{figure}
    \centering
    \includegraphics[width=.9\linewidth]{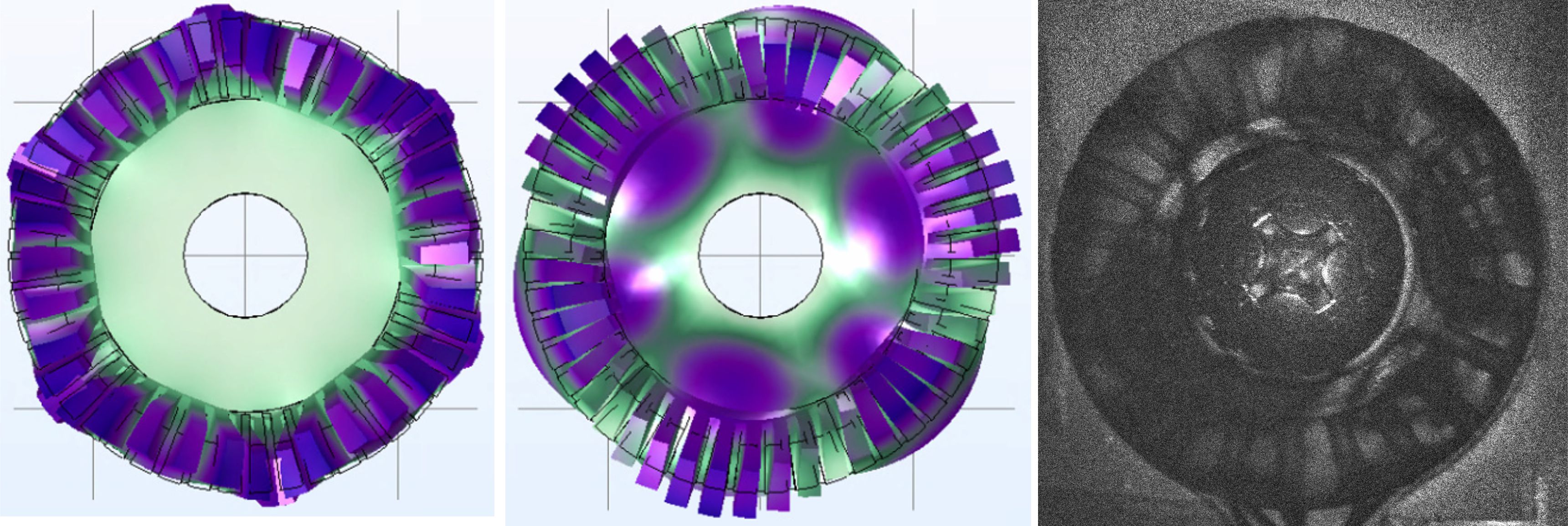}
    \caption{(Left) Simulation of out-of-plane resonate frequency at 41.154KHz. (Middle) Simulation of lateral resonate frequency at 42.757KHz. (Right) One of the notched stators performed a mixed pattern. Same as Fig. \ref{fig:notch_holo} figure (f).}
    \label{fig:MD6}
\end{figure}

The mixed pattern in Fig. \ref{fig:notch_holo} (f) needs more detailed discussion. Based on the simulation of the eigenfrequency study, out-of-plane and lateral resonate frequencies were performed at 41.154KHz and 42.757KHz, which is very close. The stator performed the mixed pattern because the two exact frequencies were combined. The simulation of two frequencies can be found in Fig. \ref{fig:MD6}. The experimental observation result of the mixed pattern is also shown in Fig. \ref{fig:MD6}, under the frequency of 42.124kHz, and it was in the middle of simulation eigenfrequencies of out-of-plane and lateral resonate frequencies.

\section{Results Analysis}

\begin{figure}
    \centering
    \includegraphics[width=0.75\linewidth]{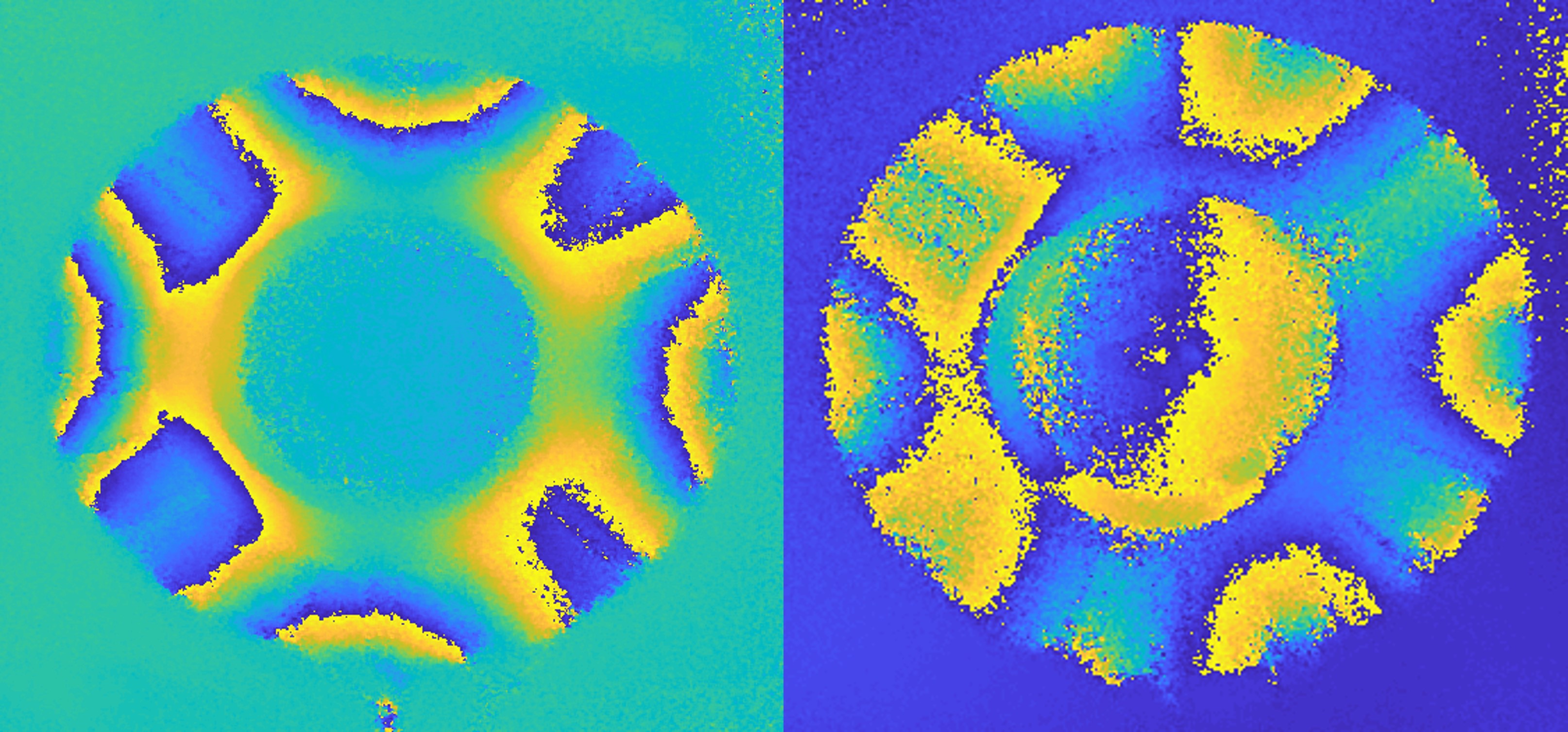}
    \caption{Stroboscopic pattern of one notched stator at $4^{th}$ (left) and $5^{th}$ (right) excitation mode}
    \label{fig:4-5holopattern}
\end{figure}

\begin{figure}
    \centering
    \includegraphics[width=1\linewidth]{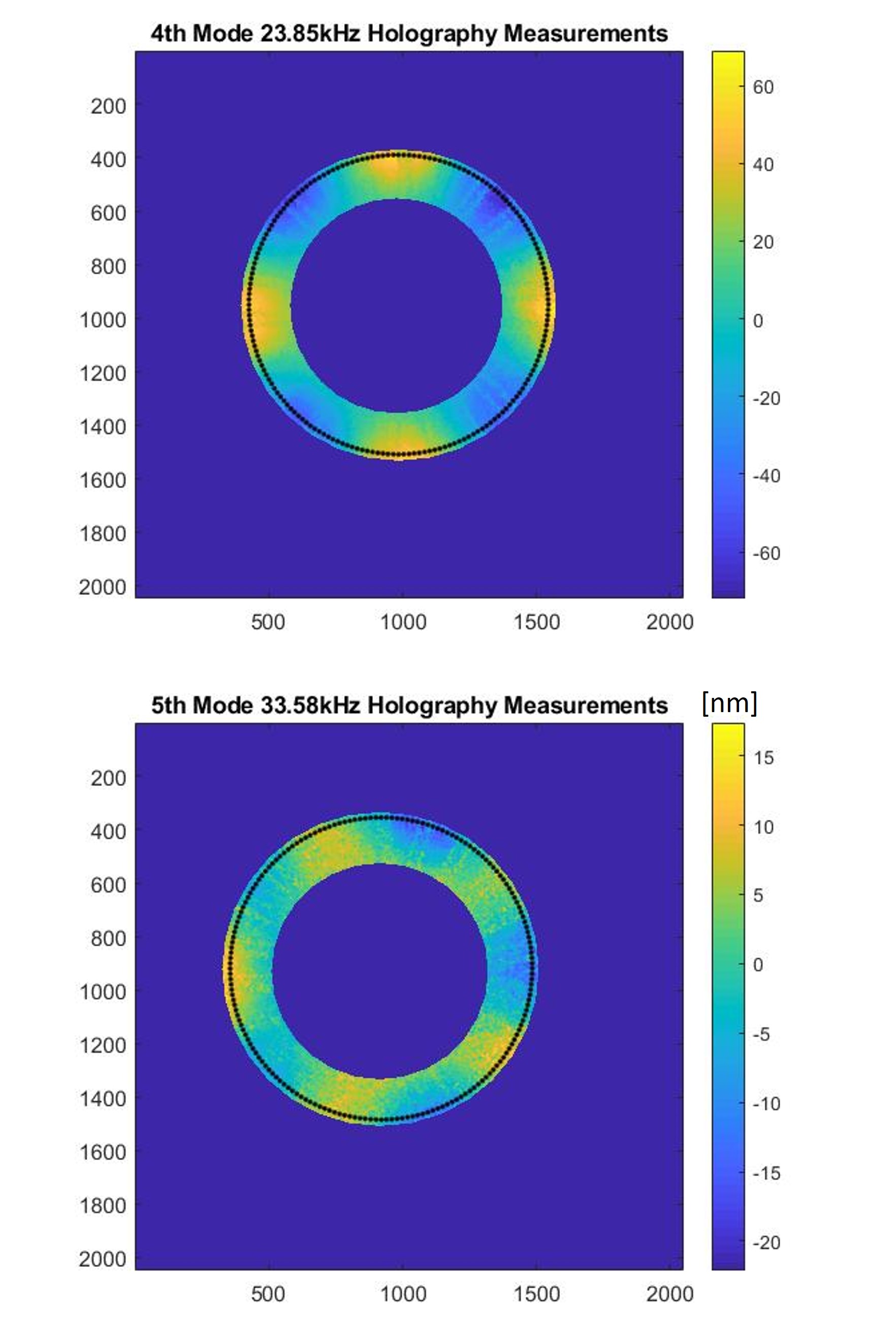}
    \caption{A 60$^\circ$ strobe phase offset with $4^{th}$, and a 15$^\circ$ strobe phase offset with $5^{th}$ respectively, for notched stator with a circle drawn along edge circumference.}
    \label{fig:stobeholo}
\end{figure}

\begin{figure}
    \centering
    \includegraphics[width=1\linewidth]{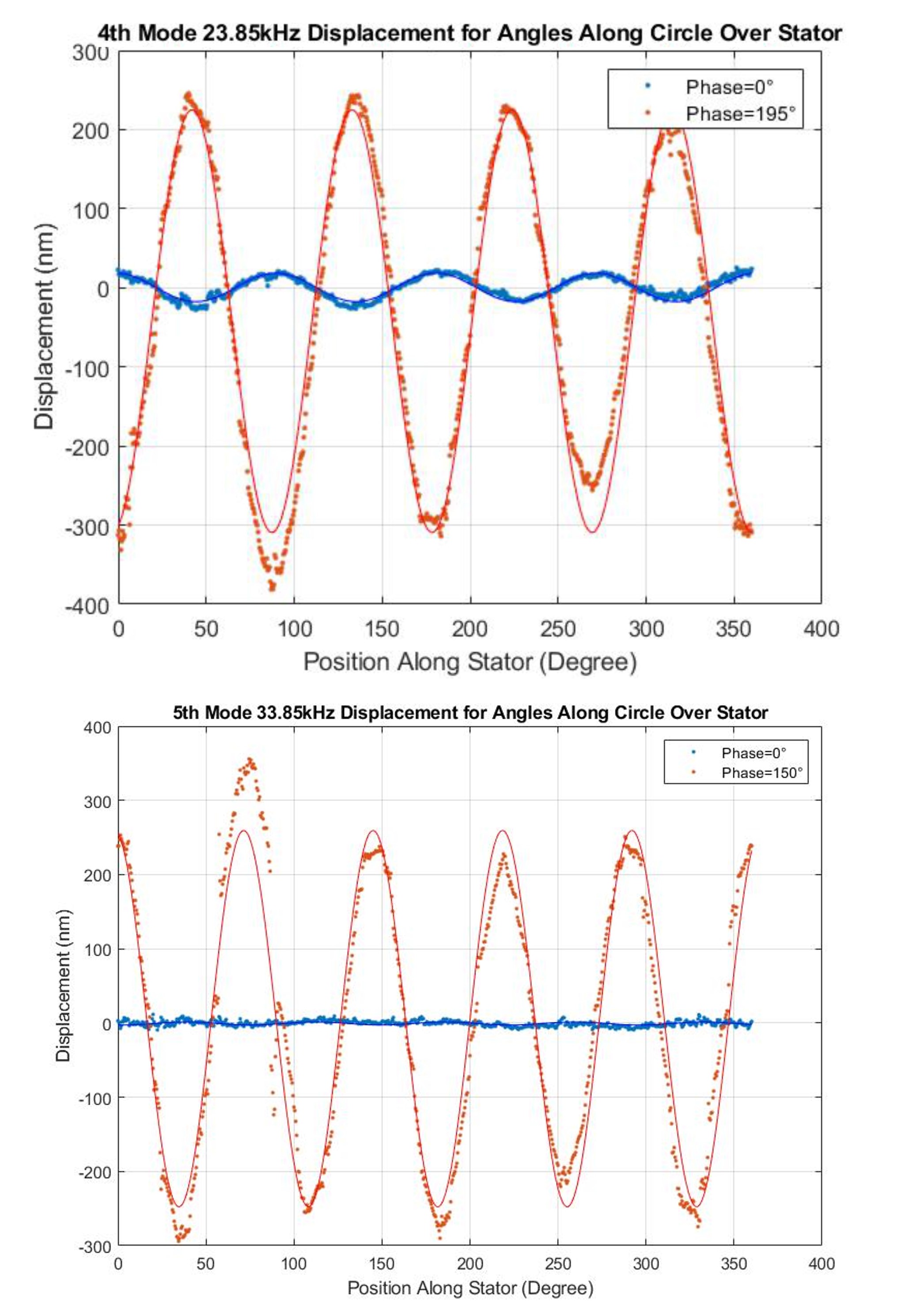}
    \caption{Least square fitting of Equ. \ref{equ:lsf} to raw data for strobe phases 0$^\circ$ and 190$^\circ$ with $4^{th}$ mode, 0$^\circ$ and 150$^\circ$ with $5^{th}$ mode respectively.}
    \label{fig:LSF}
\end{figure}

Fig. \ref{fig:4-5holopattern} shows the stroboscopic pattern of one notched stator at $4^{th}$ and $5^{th}$ excitation mode. From the patterns, we can observe the notched stator was generating traveling waves pattern symmetrically in $4^{th}$ mode. However the $5^{th}$ mode showed some asymmetry in pattern, this is likely caused by the manufacturing defects, which will affect motor performance. Quantitative observation of the motion of the peaks in the stator during its deformation is calculated by collecting the displacement values along a circle on the edge of the stator's teeth as in Fig. \ref{fig:stobeholo}, and it shows a 60$^\circ$ strobe phase offset with $4^{th}$ and a 15$^\circ$ strobe phase offset with $5^{th}$ respectively. The resulting intensity values are shown in Fig. \ref{fig:LSF}. The $5^{th}$ mode showed some asymmetry in the displacement trajectory of a specific phase compared with $4^{th}$ mode, which matched the pattern in Fig. \ref{fig:stobeholo}. 

Note that the number of peaks corresponds to the mode of the stator. We performed a least-square fit using Equ. \ref{equ:lsf}, where A is amplitude, n is the mode number, $\theta$ is angle along the drawn circle, $\phi$ is relative rotation, and $\delta$ is offset. All the results show the notched stator performed similar deformations of amplitude up to $\pm$250nm with 100V peak-to-peak voltage.

\begin{equation}
    f=Asin(n\theta + \phi) + \delta
    \label{equ:lsf}
\end{equation}

\section{Conclusion}

In this study, we discussed a modified stator design and validated it with holography interferometry technology using DHS, the notched design further proved the motion ability by using the plastic material. The traveling waves with the symmetric pattern were observed, and more importantly at the macro-level, the modified notch stator was able to rotate the rotor at a very low and non-continuous speed when testing the performance by driving a disc and performed better motion results compared to the preliminary flat design in our previous work.

\bibliographystyle{IEEEtran}
\bibliography{IEEEtran/Main}

\end{document}